\documentclass[11pt,a4paper]{article}
\usepackage[hyperref]{acl2021}
\usepackage{times}
\usepackage{latexsym}
\usepackage{bm}
\usepackage{graphicx}
\usepackage{booktabs}
\usepackage{amsmath}
\usepackage{amsfonts}
\usepackage{amssymb}
\usepackage{enumitem}
\usepackage{breakurl}

\usepackage{microtype}

\aclfinalcopy 

\usepackage{pifont}

\newcommand{\stitle}[1]{\vspace{0.3em}\noindent{\bf #1}}

\usepackage{cleveref}
\crefformat{section}{Sec.~#2#1#3}
\crefformat{subsection}{Sec.~#2#1#3}
\crefformat{subsubsection}{Sec.~#2#1#3}
\crefrangeformat{section}{Sec.#3#1#4 to~#5#2#6}
\crefrangeformat{subsection}{Sec.#3#1#4 to~#5#2#6}
\crefmultiformat{section}{Sec.#2#1#3}{ and~#2#1#3}{, #2#1#3}{ and~#2#1#3}
\usepackage{refstyle}
\Crefformat{figure}{#2Fig.~#1#3}
\Crefmultiformat{figure}{Figs.~#2#1#3}{ and~#2#1#3}{, #2#1#3}{ and~#2#1#3}
\Crefformat{table}{#2Tab.~#1#3}
\Crefmultiformat{table}{Tabs.~#2#1#3}{ and~#2#1#3}{, #2#1#3}{ and~#2#1#3}
\Crefformat{appendix}{Appx.~#2#1#3}
\crefformat{algorithm}{Alg.~#2#1#3}

\Crefformat{equation}{Eq.~#2#1#3}

\title{An Improved Baseline for Sentence-level Relation Extraction}

\author{Wenxuan Zhou \\
  University of Southern California \\
  \texttt{zhouwenx@usc.edu} \\\And
  Muhao Chen \\
  University of Southern California \\
  \texttt{muhaoche@usc.edu} \\}

\date{}

\begin{document}
\maketitle

\begin{abstract}
Sentence-level relation extraction (RE) aims at identifying the relationship between two entities in a sentence.
Many efforts have been devoted to this problem, while the best performing methods are still far from perfect.
In this paper, we revisit two problems that affect the performance of existing RE models, namely \textsc{entity representation} and \textsc{noisy or ill-defined labels}.
Our improved RE baseline, incorporated with entity representations with typed markers, achieves an $F_1$ of $74.6\%$ on TACRED, significantly outperforms previous SOTA methods.
Furthermore, the presented new baseline achieves an $F_1$ of $91.1\%$ on the refined Re-TACRED dataset, demonstrating that the pretrained language models~(PLMs) achieve high performance on this task.
We release our code\footnote{\burl{https://github.com/wzhouad/RE_improved_baseline}} to the community for future research.

\end{abstract}

\section{Introduction}

As one of the fundamental information extraction (IE) tasks,
relation extraction~(RE) aims at identifying the relationship(s) between two entities in a given piece of text from a pre-defined set of relationships of interest.
For example, given the sentence ``Bill Gates founded Microsoft together with his friend Paul Allen in 1975'' and an entity pair (``Bill Gates'', ``Microsoft''), the RE model is expected to predict the relation \texttt{ORG:FOUNDED\_BY}.
On this task, SOTA models based on PLMs \cite{devlin-etal-2019-bert,joshi-etal-2020-spanbert} have gained significant success.

Recent work on sentence-level RE can be divided into two lines.
One focuses on injecting external knowledge into PLMs.
Methods of such, including ERNIE~\cite{zhang-etal-2019-ernie} and KnowBERT~\cite{peters-etal-2019-knowledge}, take entity embedding pretrained from knowledge graphs as inputs to the Transformer.
Similarly, K-Adapter~\cite{wang2020k} introduces a plug-in neural adaptor that injects factual and linguistic knowledge into the language model.
LUKE~\cite{yamada-etal-2020-luke} further extends the pretraining objective of masked language modeling to entities and proposes an entity-aware self-attention mechanism.
The other line of work focuses on continually pretraining PLMs on text with linked entities using relation-oriented objectives.
Specifically, BERT-MTB~\cite{baldini-soares-etal-2019-matching} proposes a matching-the-blanks objective that decides whether two relation instances share the same entities.
Despite extensively studied, existing RE models still perform far from perfect.
On the commonly-used benchmark TACRED~\cite{zhang-etal-2017-position}, the SOTA $F_1$ result only increases from $70.1\%$~(BERT$_{\text{LARGE}}$) to $72.7\%$~(LUKE) after applying PLMs to this task.
It is unclear what building block is missing to constitute a promising RE system. 

In this work, we discuss two obstacles that have hindered the performance of existing RE models.
First, the RE task provides a structured input of both the raw texts and \emph{side information} of the entities, such as entity names, spans, and types (typically provided by NER models), which are shown important to the performance of RE models~\cite{peng-etal-2020-learning}.
However, existing methods fall short of representing the entity information comprehensively in the text, leading to limited characterization of the entities.
Second, human-labeled RE datasets~(e.g., TACRED), may contain a large portion of noisy or ill-defined labels, causing the model performance to be misestimated.
\citet{alt-etal-2020-tacred} relabeled the development and test set of TACRED and found that $6.62\%$ of labels are incorrect.
\citet{stoica2021re} refined many ill-defined relation types and further re-annotated the TACRED dataset using an improved annotation strategy to ensure high-quality labels.
To this end, we propose an improved RE baseline, where we introduce the typed entity marker to sentence-level RE, which leads to promising improvement of performance over existing RE models.

We evaluate our model on TACRED~\cite{zhang-etal-2017-position}, TACREV~\cite{alt-etal-2020-tacred}, and Re-TACRED~\cite{stoica2021re}.
Using RoBERTa~\cite{liu2019roberta} as the backbone, our improved baseline model achieves an $F_1$ of $74.6\%$ and $83.2\%$ on TACRED and TACREV, respectively, significantly outperforming various SOTA RE models.
Particularly, our baseline model achieves an $F_1$ of $91.1\%$ on Re-TACRED, demonstrating that PLMs can achieve much better results on RE than shown in previous work.\footnote{This work first appeared as a technical report on arXiv in Feb 2021 \cite{zhou2021improved}.
Since then, the proposed techniques have been incorporated into several follow-up works \cite{chen2022knowprompt,wang-etal-2022-rely,wang-etal-2022-graphcache,lu2022summarization,han2021ptr,kulkarni-etal-2022-learning} that are published before this version of the paper.
}

\section{Method}

In this section, we first formally define the relation extraction task in \Cref{sec:task_definition}, and then present our model architecture and entity representation techniques in \Cref{sec:model_architecture} and \Cref{sec:entity_representation}.

\subsection{Problem Definition}
\label{sec:task_definition}
In this paper, we focus on sentence-level RE.
Specifically, given a sentence $\bm{x}$ mentioning an entity pair $(e_s, e_o)$, referred as the subject and object entities, respectively, the task of sentence-level RE is to predict the relationship $r$ between $e_s$ and $e_o$ from $\mathcal{R} \cup \{\textsc{NA}\}$, where $\mathcal{R}$ is a pre-defined set of relationships of interest.
If the text does not express any relation from $\mathcal{R}$, the entity pair will be accordingly labeled \textsc{NA}.

\subsection{Model Architecture}
\label{sec:model_architecture}
Our RE classifier is an extension of 
previous PLM-based RE models~\cite{baldini-soares-etal-2019-matching}.
Given the input sentence $\bm{x}$, we first mark the entity spans and entity types using techniques presented in~\Cref{sec:entity_representation}, then feed the processed sentence into a PLM to get its contextual embedding.
Finally, we feed the hidden states of the subject and object entities in the language model's last layer, i.e., $\bm{h}_\text{subj}$ and $\bm{h}_\text{obj}$, into the softmax classifier:
\begin{align*}
    \bm{z} &= \text{ReLU}\left(\bm{W}_\text{proj}\left[\bm{h}_\text{subj}, \bm{h}_\text{obj}\right]\right), \\
    \mathrm{P}(r) &= \frac{\exp(\bm{W}_r \bm{z} + \bm{b}_{r})}{\sum_{r'\in \mathcal{R} \cup \{\textsc{NA}\}} \exp(\bm{W}_{r'} \bm{z}  + \bm{b}_{r'})},
\end{align*}
where $\bm{W}_\text{proj} \in \mathbb{R}^{2d \times d}$, $\bm{W}_{r}, \bm{W}_{{r'}} \in \mathbb{R}^d, \bm{b}_{r}, \bm{b}_{r'} \in \mathbb{R}$ are model parameters.
In inference, the classifier returns the relationship with the maximum probability as the predicted relationship.
\subsection{Entity Representation}
\label{sec:entity_representation}
For sentence-level RE, the names, spans, and NER types of subject and object entities are provided in the structured input.
Such composite entity information provides useful clues to the relation types.
For example, the relationship \texttt{ORG:FOUNDED\_BY} is more likely to hold when entity types of subject and object are \texttt{ORGANIZATION} and \texttt{PERSON}, respectively, and is less likely for instances where the entity types do not match.
The entity information needs to be represented in the input text, allowing it to be captured by the PLMs.
Such techniques have been studied in previous work~\cite{zhang-etal-2017-position,baldini-soares-etal-2019-matching,wang2020k}, while many of them fall short of capturing all types of the provided information.
In this paper, we re-evaluate existing entity representation techniques and also seek to propose a better one.
We evaluate the following techniques:


\begin{itemize}[leftmargin=1em]
    \setlength\itemsep{0em}

    \item \textbf{Entity mask}~\cite{zhang-etal-2017-position}. This technique introduces new special tokens \texttt{[SUBJ-\emph{TYPE}]} or \texttt{[OBJ-\emph{TYPE}]} to mask the subject or object entities in the original text, where \texttt{\emph{TYPE}} is substituted with the respective entity type.
    This technique was originally proposed in the PA-LSTM model~\cite{zhang-etal-2017-position}, and was later adopted by PLMs such as SpanBERT~\cite{joshi-etal-2020-spanbert}.
    \citet{zhang-etal-2017-position} claim that this technique prevents the RE model from over-fitting specific entity names, leading to more generalizable inference.

    \item \textbf{Entity marker}~\cite{zhang-etal-2019-ernie,baldini-soares-etal-2019-matching}. This technique introduces special tokens pairs \texttt{[E1]}, \texttt{[/E1]} and \texttt{[E2]}, \texttt{[/E2]} to enclose the subject and object entities, therefore modifying the input text to the format of ``\texttt{[E1]} \textsc{subj} \texttt{[/E1]} ... \texttt{[E2]} \textsc{obj} \texttt{[/E2]}''\footnote{\textsc{subj} and \textsc{obj} are respectively the original token spans of subject and object entities.}.

    \item \textbf{Entity marker (punct)}~\cite{wang2020k,zhou2021atlop}. This technique is a variant of the previous technique that encloses entity spans using punctuation.
    It modifies the input text to ``@ \textsc{subj} @ ... \# \textsc{obj} \#''. The main difference from the previous technique is that this one does not introduce new special tokens into the model's reserved vocabulary.

    \item \textbf{Typed entity marker}~\cite{zhong2020frustratingly}. This technique further incorporates the NER types into entity markers.
    It introduces new special tokens \texttt{$\langle$S:\emph{TYPE}$\rangle$}, \texttt{$\langle$/S:\emph{TYPE}$\rangle$}, \texttt{$\langle$O:\emph{TYPE}$\rangle$}, \texttt{$\langle$/O:\emph{TYPE}$\rangle$}, where \texttt{\emph{TYPE}} is the corresponding NER type given by a named entity tagger. The input text is accordingly modified to ``\texttt{$\langle$S:\emph{TYPE}$\rangle$} \textsc{subj} \texttt{$\langle$/S:\emph{TYPE}$\rangle$} ... \texttt{$\langle$O:\emph{TYPE}$\rangle$} \textsc{obj} \texttt{$\langle$/O:\emph{TYPE}$\rangle$}''.

    \item \textbf{Typed entity marker (punct)}. We propose a variant of the typed entity marker technique that marks the entity span and entity types without introducing new special tokens.
    This is to enclose the subject and object entities with ``@'' and ``\#'', respectively.
    We also represent the subject and object entity types using their label text, which is prepended to the entity spans and is enclosed by ``*'' for subjects or ``$\wedge$'' for objects.
    The modified text is ``@ * \textsc{\textit{subj-type}} * \textsc{subj} @ ... \# $\wedge$ \textsc{\textit{obj-type}} $\wedge$ \textsc{obj} \# '', where \textsc{\textit{subj-type}} and \textsc{\textit{obj-type}} is the label text of NER types.
\end{itemize}

The embedding of all new special tokens is randomly initialized and updated during fine-tuning.

\section{Experiments}

\begin{table*}[!t]
\centering
\scalebox{0.76}{
    \begin{tabular}{p{4.3cm}p{8.2cm}ccc}
         \toprule
         Method& Input Example& BERT$_{\text{BASE}}$& BERT$_{\text{LARGE}}$& RoBERTa$_{\text{LARGE}}$ \\
         \midrule
         Entity mask& \texttt{[SUBJ-PERSON]} \text{was born in} \texttt{[OBJ-CITY]}\text{.}& 69.6& 70.6& 60.9 \\
         Entity marker& \texttt{[E1]} \text{Bill} \texttt{[/E1]} \text{was born in} \texttt{[E2]} \text{Seattle} \texttt{[/E2]}\text{.}& 68.4& 69.7& 70.7 \\
         Entity marker (punct)& \text{@ Bill @ was born in \# Seattle \#.}& 68.7& 69.8& 71.4 \\
         Typed entity marker& \texttt{$\langle$S:PERSON$\rangle$} \text{Bill} \texttt{$\langle$/S:PERSON$\rangle$} \text{was born in} \texttt{$\langle$O:CITY$\rangle$} \text{Seattle} \texttt{$\langle$/O:CITY$\rangle$}\text{.}& \textbf{71.5}& \textbf{72.9}& 71.0 \\
         Typed entity marker (punct)& \text{@ * person * Bill @ was born in \# $\wedge$ city $\wedge$ Seattle \#.}& 70.9& 72.7& \textbf{74.6} \\
         \bottomrule
    \end{tabular}}
    \caption{Test $\bm{F_1}$ (in \%) of different entity representation techniques on TACRED. For each technique, we also provide the processed input of an example text \textit{``Bill was born in Seattle''}. Typed entity markers (original and punct) significantly outperforms others.}
    \label{tab::representation}
\end{table*}

In this section, we evaluate the proposed techniques based on widely used RE benchmarks. The evaluation starts by first identifying the best-performing entity representation technique (\Cref{ssec:exp_rep}), which is further incorporated into our improved RE baseline to be compared against prior SOTA methods (\Cref{ssec:main}).
Due to space limits, we study in the Appendix of how the incorporated techniques lead to varied generalizability on unseen entities (\Cref{ssec:exp_unseen}) and how they perform under annotation errors (\Cref{ssec:noise}).

\subsection{Preliminaries}\label{ssec:exp_prelim}

\stitle{Datasets.} The datasets we have used in the experiments include three versions of TACRED: the original TACRED~\cite{zhang-etal-2017-position}, TACREV~\cite{alt-etal-2020-tacred}, and Re-TACRED~\cite{stoica2021re}.
\citet{alt-etal-2020-tacred} observed that the TACRED dataset contains about $6.62\%$ noisily-labeled instances and relabeled the development and test set.
\citet{stoica2021re} further refined the label definitions in TACRED and relabeled the whole dataset.
We provide the statistics of the datasets in~\Cref{ssec:data_statistics}.

\stitle{Compared methods.}
We compare with the following methods.
\textbf{PA-LSTM}~\cite{zhang-etal-2017-position} adopts bi-directional LSTM~\cite{Hochreiter1997LongSM} and positional-aware attention~\cite{Bahdanau2015NeuralMT} to encode the text into an embedding, which is then fed into a softmax layer to predict the relation.
\textbf{C-GCN}~\cite{zhang-etal-2018-graph} is a graph-based model, which feeds the pruned dependency tree of the sentence into the graph convolutional network~\cite{Kipf2017SemiSupervisedCW} to obtain the representation of entities.
\textbf{SpanBERT}~\cite{joshi-etal-2020-spanbert} is a PLM based on the Transformer~\cite{vaswani2017attention}. It extends BERT~\cite{devlin-etal-2019-bert} by incorporating a training objective of span prediction and achieves improved performance on RE.
\textbf{KnowBERT}~\cite{peters-etal-2019-knowledge} jointly trains a language model and an entity linker, which allows the subtokens to attend to entity embedding that is pretrained on knowledge bases.
\textbf{LUKE}~\cite{yamada-etal-2020-luke} pretrains the language model on both large text corpora and knowledge graphs. It adds frequent entities into the vocabulary and proposes an entity-aware self-attention mechanism.

\stitle{Model configurations.}
For the compared methods, we rerun their officially released code using the recommended hyperparameters in their papers.
Our model is implemented based on HuggingFace's Transformers~\cite{wolf-etal-2020-transformers}.
Our model is optimized with Adam~\cite{Kingma2015AdamAM} using the learning rate of $5\mathrm{e}{-5}$ on BERT$_\text{BASE}$, and $3\mathrm{e}{-5}$ on BERT$_\text{LARGE}$ and RoBERTa$_\text{LARGE}$, with a linear warm-up~\cite{Goyal2017AccurateLM} of for the first $10\%$ steps followed by a linear learning rate decay to 0.
We use a batch size of 64 and fine-tune the model for 5 epochs on all datasets.
For all experiments, we report the median $F_1$ of 5 runs of training using different random seeds.

\subsection{Analysis on Entity Representation}\label{ssec:exp_rep}

We first provide an analysis on different entity representation techniques. In this analysis, we use the base and large versions of BERT~\cite{devlin-etal-2019-bert} and the large version of RoBERTa~\cite{liu2019roberta} as the encoder.
\Cref{tab::representation} shows the performance of the PLMs incorporated with different entity representation techniques.
For each technique, we also provide an example of the processed text.
We have several observations from the results.
First, the typed entity marker and its variants outperform untyped entity representation techniques by a notable margin.
Especially, the RoBERTa model achieves an $F_1$ score of $74.6\%$ using the typed entity marker (punct), which is significantly higher than the SOTA result of $72.7\%$ by LUKE~\cite{yamada-etal-2020-luke}.
This shows that representing all categories of entity information is helpful to the RE task.
It also shows that keeping entity names in the input improves the performance of RE models.
Second, symbols used in entity markers have an obvious impact on the performance of RE models.
Although the original and \emph{punct} versions of entity representation techniques represent the same categories of entity information, they do lead to a difference in model performance.
Particularly, introducing new special tokens hinders the model performance drastically on RoBERTa.
On RoBERTa$_\text{LARGE}$, the entity marker underperforms the entity marker (punct) by $0.7\%$, the typed entity marker underperforms the typed entity marker (punct) by $3.6\%$, while the entity mask gets a much worse result of $60.9\%$.

\subsection{Comparison with Prior Methods}\label{ssec:main}

The prior experiment has found RoBERTa$_{\text{LARGE}}$ with the typed entity marker (punct) to be the best-performing RE model.
We now compare our improved baseline with methods in prior studies.

The experimental results are shown in \Cref{tab::tacred}.
We evaluate all methods on TACRED, TACREV, and Re-TACRED.
Incorporated with the typed entity marker (punct) and using RoBERTa$_{\text{LARGE}}$ as the backbone, our improved baseline model achieves new SOTA results over previous methods on all datasets.
However, we observe that on Re-TACRED, the gain from the typed entity marker is much smaller compared to TACRED and TACREV, decreasing from $3.1-3.9\%$ and $2.0-3.4\%$ to $0.2-0.8\%$ of $F_1$.
This observation could be attributed to the high noise rate in TACRED, in which the noisy labels are biased towards the side information of entities.

\begin{table}[!t]
\setlength{\tabcolsep}{1pt}
\centering
{
\scalebox{0.72}{
    \begin{tabular}{p{5cm}ccc}
         \toprule
         \textbf{Model} & \multicolumn{1}{c}{\textbf{TACRED}} & \multicolumn{1}{c}{\textbf{TACREV}}& \multicolumn{1}{c}{\textbf{Re-TACRED}} \\
          &Test $F_1$ & Test $F_1$ & Test $F_1$ \\
         \midrule
         \textit{Sequence-based Models} \\
         PA-LSTM~\cite{zhang-etal-2017-position}& 65.1& 73.3$^\ddagger$ & 79.4$^\dagger$ \\
         C-GCN~\cite{zhang-etal-2018-graph}& 66.3& 74.6$^\ddagger$ &80.3$^\dagger$ \\
         \midrule
         \textit{Transformer-based Models} \\
         BERT$_{\text{BASE}}$ + entity marker& 68.4& 77.2& 87.7 \\
         BERT$_{\text{LARGE}}$ + entity marker& 69.7& 77.9& 89.2  \\
         RoBERTa$_{\text{LARGE}}$ + entity marker& 70.7& 81.2& 90.5\\
         SpanBERT~\cite{joshi-etal-2020-spanbert}& 70.8& 78.0\rlap{$^*$}& 85.3\rlap{$^\dagger$} \\
         KnowBERT~\cite{peters-etal-2019-knowledge}& 71.5& 79.3\rlap{$^*$}& - \\
         LUKE~\cite{yamada-etal-2020-luke}& 72.7& 80.6\rlap{$^\ddagger$}& 90.3\rlap{$^\ddagger$} \\
         \midrule
         \textit{Improved RE baseline}\\
         BERT$_{\text{BASE}}$ + typed entity marker& 71.5 & 79.3 & 87.9 \\
         BERT$_{\text{LARGE}}$ + typed entity marker & 72.9 & 81.3 & 89.7 \\
         RoBERTa$_{\text{LARGE}}$ + typed entity marker (punct)& \textbf{74.6}& \textbf{83.2}& \textbf{91.1} \\
         \bottomrule
    \end{tabular}}}
    \caption{$F_1$ (in \%) on the test sets. * marks re-implemented results from \citet{alt-etal-2020-tacred}. $\dagger$ marks re-implemented results from \citet{stoica2021re}. $\ddagger$ marks our re-implemented results.
    }
    \label{tab::tacred}
\end{table}

To assess how the presented techniques contribute to robustness and generalizability of RE, we provide more analyses on
varied generalizability on unseen entities (\Cref{ssec:exp_unseen}) and the performance under annotation errors (\Cref{ssec:noise}) in the Appendix.

\section{Conclusion}
In this paper, we present a simple yet strong RE baseline that offers new SOTA performance, along with a comprehensive study to understand its prediction generalizability and robustness.
Specifically, we revisit two technical problems in sentence-level RE, namely \emph{entity representation} and \emph{noisy or ill-defined labels}.
We propose an improved entity representation technique, which significantly outperforms existing sentence-level RE models.
Especially, our improved RE baseline achieves an $F_1$ score of $91.1\%$ on the Re-TACRED dataset, showing that PLMs already achieve satisfactory performance on this task.
We hope the proposed techniques and analyses can benefit future research on RE.


\section*{Acknowledgement}

We appreciate the reviewers for their insightful comments and suggestions.
This work supported by the National Science Foundation of United States Grant IIS 2105329, and a Cisco Research Award.

\bibliographystyle{acl_natbib}
\bibliography{acl2021}

\clearpage
\appendix

\section{Dataset Statistics}
\label{ssec:data_statistics}
\begin{table}[!h]
    \centering
    \scalebox{0.76}{
    \begin{tabular}{p{2.4cm}cccc}
    \toprule
    \textbf{Dataset}& \# train& \# dev& \# test& \# classes\\
    \midrule
    TACRED& 68124& 22631& 15509& 42\\
    TACREV& 68124& 22631& 15509& 42\\
    Re-TACRED& 58465& 19584& 13418& 40\\
    \bottomrule
    \end{tabular}}
    \caption{Statistics of datasets.}
    \label{tab:data_statistics}
\end{table}
The statistics of the datasets are shown in~\Cref{tab:data_statistics}.

\section{Analysis on Unseen Entities}\label{ssec:exp_unseen}

Some previous work~\cite{zhang-etal-2018-graph,joshi-etal-2020-spanbert} claims that entity names may leak superficial clues of the relation types, allowing heuristics to hack the benchmark.
They show that neural RE models can achieve high evaluation results only based on the subject and object entity names even without putting them into the original sentence.
They also suggest that RE models trained without entity masks may not generalize well to unseen entities.
However, as the provided NER types in RE datasets are usually coarse-grained, using entity masks may lose the meaningful information of entities.
Using entity masks also contradicts later studies' advocacy of injecting entity knowledge into RE models~\cite{zhang-etal-2019-ernie,peters-etal-2019-knowledge,wang2020k}.
If RE models should not consider entity names, it is unreasonable to suppose that they can be improved by external knowledge graphs.

To evaluate whether the RE model trained without entity mask can generalize to unseen entities, we propose a \textit{filtered} evaluation setting. Specifically, we remove all test instances containing entities from the training set of TACRED, TACREV, and Re-TACRED. This results in \textit{filtered test sets} of 4,599 instances on TACRED and TACREV, and 3815 instances on Re-TACRED.
These filtered test sets only contain instances with unseen entities during training.

We present the evaluation results on the filtered test set in \Cref{tab::filter}.
We compare the performance of models with entity mask or typed entity marker representations, between which the only difference lies in whether to include entity names in entity representations or not.
Note that as the label distributions of the original and filtered test set are different, their results are not directly comparable.
Still, the \emph{typed entity marker} consistently outperforms the \emph{entity mask} on all encoders and datasets, which shows that RE models can learn from entity names and generalize to unseen entities.
Our finding is consistent with \citet{peng-etal-2020-learning}, whose work suggests that entity names can provide semantically richer information than entity types to improve the RE model.

\begin{table}[!t]
\centering
{
\scalebox{0.72}{
 \setlength{\tabcolsep}{1pt}
    \begin{tabular}{p{5cm}ccc}
         \toprule
         \textbf{Model} & \multicolumn{1}{c}{\textbf{TACRED}} & \multicolumn{1}{c}{\textbf{TACREV}}& \multicolumn{1}{c}{\textbf{Re-TACRED}} \\
          &Test $F_1$ & Test $F_1$ & Test $F_1$ \\
         \midrule
         BERT$_{\text{BASE}}$ + entity mask& 75.2& 82.7& 83.8 \\
         BERT$_{\text{BASE}}$ + typed entity marker& 75.8 & 83.7 & 87.0 \\
         \midrule
         BERT$_{\text{LARGE}}$ + entity mask& 75.8& 83.7& 85.6  \\
         BERT$_{\text{LARGE}}$ + typed entity marker & 77.0 & 85.3 & 89.8 \\
         \midrule
         RoBERTa$_{\text{LARGE}}$ + entity mask& 69.4& 78.8& 82.2\\
         RoBERTa$_{\text{LARGE}}$ + typed entity marker (punct)& 78.7& 86.9& 91.7 \\
         \bottomrule
    \end{tabular}}}
    \caption{Test $F_1$ on the filtered test sets. The typed entity marker consistently outperforms the entity mask, showing that knowledge from entity names can generalize to unseen entities.}
    \label{tab::filter}
\end{table}

\begin{table}[!t]
\centering
{
\scalebox{0.72}{
 \setlength{\tabcolsep}{1pt}
    \begin{tabular}{p{4cm}ccc}
         \toprule
         \textbf{Model} & BERT$_{\text{BASE}}$& BERT$_{\text{LARGE}}$& RoBERTa$_{\text{LARGE}}$ \\
         \midrule
         Entity marker& 83.8& 86.0& 88.6 \\
         Typed entity marker (punct for RoBERTa)& 84.3 & 87.5 & 89.4 \\
         Gain& +0.5& +1.5& +0.8 \\
         \midrule
         Gain on TACRED& +3.1& +3.2& +3.9  \\
         Gain on TACREV& +2.1& +3.4& +2.0 \\
         \bottomrule
    \end{tabular}}}
    \caption{Test $F_1$ on the clean test set of TACRED. The gain on the clean test set is smaller than on TACRED and TACREV.}
    \label{tab::clean}
\end{table}

\section{Analysis on Annotation Errors}\label{ssec:noise}
Our model achieves a smaller performance gain on Re-TACRED compared to TACRED and TACREV.
We find that this difference can be mainly attributed to the annotation errors in their evaluation sets.
Specifically, we create a clean TACRED test set by pruning all instances in the TACRED test set, of which the annotated relation is different in the Re-TACRED test set.
The remaining instances are considered clean.
Note that as the label sets of TACRED and Re-TACRED are different, instances of some classes cannot be found in Re-TACRED and are thus completely pruned.
We train the model on the original~(noisy) training set and show the results on the clean test set in~\Cref{tab::clean}.
We observe a similar drop of performance gain on the clean TACRED test set.
It shows that the annotation errors in TACRED and TACREV can lead to overestimation of the performance of models depending on the side information of entities.
We hypothesize that in data annotation, much noise may be created as some annotators label the relation only based on the two entities without reading the whole sentence.
Therefore, integrating NER types into the entity representation can brings larger performance gain.
Overall, this experiment shows that the evaluation sets of both TACRED and TACREV are biased and unreliable.
We recommend future work on sentence-level RE should use Re-TACRED as the evaluation benchmark.

\end{document}